\documentclass[10pt,twocolumn,letterpaper]{article}

\usepackage{iccv}
\usepackage{times}
\usepackage{epsfig}
\usepackage{graphicx}
\usepackage{amsmath}
\usepackage{amssymb}
\usepackage{booktabs}

\usepackage[english]{babel}
\usepackage[utf8]{inputenc}
\usepackage[ruled,linesnumbered]{algorithm2e}

\usepackage{comment}
\usepackage{multirow}

\usepackage[pagebackref=true,breaklinks=true,letterpaper=true,colorlinks,bookmarks=false]{hyperref}

\newcommand{\AG}[1]{\textcolor{blue}{\bf [AG: #1]}}

\newcommand{\BR}[1]{\textcolor{red}{\bf [BR: #1]}}

\DeclareMathOperator*{\IoU}{IoU}

\iccvfinalcopy 

\def\iccvPaperID{21} 

\ificcvfinal\fi

\begin{document}

\title{Temporal Coherence for Active Learning in Videos}

\author{Javad Zolfaghari Bengar$^{1,2}$\\
\and
Abel Gonzalez-Garcia$^1$ \\
\and
Gabriel Villalonga$^{1,2}$ \\
\and
Bogdan Raducanu$^{1,2}$\\
\and
Hamed H. Aghdam$^1$\\
\and 
Mikhail Mozerov$^{1,2}$\\
\and
Antonio M. L\'{o}pez$^{1,2}$\\
\and
Joost van de Weijer$^{1,2}$\\
Computer Vision Center (CVC)$^1$, Univ. Aut\`{o}noma of Barcelona (UAB)$^2$\\
{\tt\small \{jzolfaghari,agonzalez,gvillalonga,bogdan,haghdam,mozerov,antonio,joost\}@cvc.uab.es}
}

\maketitle

\begin{abstract}
Autonomous driving systems require huge amounts of data to train. Manual annotation of this data is time-consuming and prohibitively expensive since it involves human resources. Therefore, active learning emerged as an alternative to ease this effort and to make data annotation more manageable. In this paper, we introduce a novel active learning approach for object detection in videos by exploiting temporal coherence. Our active learning criterion is based on the estimated number of errors in terms of false positives and false negatives. 
The detections obtained by the object detector are used to define the nodes of a graph and tracked forward and backward to temporally link the nodes. 
Minimizing an energy function defined on this graphical model provides estimates of both false positives and false negatives. Additionally, we introduce a synthetic video dataset, called SYNTHIA-AL, specially designed to evaluate active learning for video object detection in road scenes. 
Finally, we show that our approach outperforms active learning baselines tested on two datasets. 
\end{abstract}

\section{Introduction}
For autonomous driving systems, the quality of object detection is of key importance. Its progress in recent years has been notable, partially due to the presence of large datasets~\cite{geiger2013vision,yu2018bdd100k}.  However, pushing detectors to further improve and finally be close to flawless, requires the collection of ever larger labeled datasets, which is both time and labor expensive. Active learning methods~\cite{settles2009active} tackle this problem by reducing the required annotation effort. The key idea behind active learning is that a machine learning model can achieve a satisfactory performance with a subset of the training samples if it is allowed to choose which samples to label. This contrasts with passive learning, where the data to be labeled is taken at random without taking into account the potential benefit of annotating each sample. 

Active learning has been mainly investigated for the image classification task ~\cite{porikli2012scalable,xin2013cvpr,snoek2015transfer,falcao2015pr,guo2014eccv,yan2016tmm,deng2018pr}. Only few works have investigated active learning for object detection, even though the problem of active learning is more pertinent for object detection than for image classification since the labelling effort also includes the more expensive annotation of the bounding box~\cite{soatto2014cvpr}. For instance, in~\cite{vangool2012cvpr,grauman2014ijcv} the object detector is learned interactively in an incremental manner using a simple margin approach to select the most uncertain images. In~\cite{roy2018bmvc}, the active learning approach is based on a ‘query-by-committee’ strategy.

In this work we focus on active learning for object detection in videos. To the best of our knowledge, we are the first to consider this scenario. Object detection in videos has become of great interest ever since the introduction of the large-scale video object detection challenge ImageNet-VID~\cite{ImageNetVID2015}. The task has proven highly challenging due to phenomena such as detector flicker~\cite{rosenfeld2018elephant,jin2018unsupervised}, i.e.\ the drastic effects in the predicted outputs given by small changes in the images. This has spawn a multitude of video-specific approaches~\cite{kang2017object,kang2018t,zhu2017flow,zhu2017deep,wang2018fully} that require comprehensive video annotation. However, exhaustively annotating all object instances in every frame is extremely costly. Possibly because of this, recent datasets for autonomous driving~\cite{yu2018bdd100k,bdd-nexar} only offer a small subset of frames with object ground-truth annotations. 

Video data has the inherent property of \emph{temporal coherence}, i.e. nearby frames are expected to contain the same instances in nearby locations. This property can be exploited to identify frames in which the detector might have wrongly detected objects (there is no support in nearby frames) or frames in which the detector failed to detect an object (there is evidence of the object in the surrounding frames). These frames are expected to be more beneficial to annotate than others, leading to potentially more accurate models when used for training. 

In this paper, we confirm that annotating those frames that contain detection errors leads to higher accuracy given a limited annotation budget. We consider two types of errors, false positives and false negatives, and show the effect of selecting either type. This exploratory experiment suggests a potentially powerful approach for active learning. Motivated by this, we develop a novel method to estimate detection errors in videos by exploiting the temporal coherence in the videos. 
We track detections forward and backward and define a graph on the detections that are temporally linked. Minimization of an energy function defined on this graphical model provides us with the detection of false positives and false negatives. These we subsequently use to select the frames to be annotated. In summary, the contributions of this paper are: 
\begin{itemize}
    \item We propose a new method for active learning in videos which exploits the temporal coherence.
    \item We propose a new synthetic dataset specially designed for active learning in road scene videos.
    \item Our proposed method outperforms several baseline methods both on synthetic and real video data. 
\end{itemize}

\section{Related Work}
\paragraph{Active learning for object detection.}
A critical aspect for an active learner is represented by the strategy used to query the next sample to be labeled. Four main query frameworks exist, which rely mostly on heuristics: informativeness~\cite{hauptmann2015ijcv,gal2017icml, guo2010nips,gu2014modelchange}, representativeness~\cite{falcao2015pr,sener2018active}, hybrid~\cite{zhou2014hybrid,loog2018maxvariance}, and performance-based~\cite{ungar2007regression,han2012nips,wu2018kdd,loog2018regression}. Among all these, informativeness-based approaches are the most successful ones. 
A comprehensive survey of these frameworks and a detailed discussion can be found in~\cite{settles2009active}.
Active learning has been successfully applied to a series of traditional computer vision tasks, such as image classification~\cite{urtasun2007al,porikli2012scalable,snoek2015transfer} (including medical image classification  ~\cite{falcao2015pr} and scene classification~\cite{guo2014eccv}), visual question answering (VQA)~\cite{parikh2017vqa}, image retrieval~\cite{zhang2010retrieval}, remote sensing~\cite{deng2018pr}, action localization~\cite{ghanem2018eccv}, and regression~\cite{freytag2014influential,denzler2018bmvc}. 

With a strong emphasis on image classification, active learning for object detection has received less attention than expected due to the difficulty to aggregate several object hypothesis at frame level. 
Recently, ~\cite{Yoo_2019_CVPR} employed a loss module to learn the loss of a target model and select the images based on their output loss. However, in hybrid tasks such as object detection learning the loss is challenging. In ~\cite{roy2018bmvc}, the active learning approach is based on a `query-by-committee' strategy. A committee of classifiers is formed by the last convolutional layer of the base network together with the
extra convolutional layers of the SSD architecture~\cite{ssd2016eccv}. The disagreement between them for each candidate bounding box in an image is used as query strategy.
In~\cite{grauman2014ijcv}, the authors propose a system that learns object detectors on-the-fly, by refining its models via crowd-sourced annotations of web images. As active learning criterion, they use a simple margin approach which selects the most uncertain images which should be annotated. 
A similar idea is reported in~\cite{vangool2012cvpr}, where an object detector is learned interactively, in an incremental manner. The system selects the images most likely to require user input based on an estimated annotation cost computed in terms of false positive and false negative detections. Other approaches to reduce the annotation cost for object detection are based on domain adaptation~\cite{hoffman2014lsda} or transfer learning~\cite{ferrari2012cvpr}.

In the current work, we introduce a novel active learning approach for object detection in videos, which exploits the temporal coherence of the found detections. The query strategy is based on the number of false positives and false negatives detections identified using a graphical model.

\paragraph{Temporal coherence in video object detection.}
Several video object detection approaches~\cite{han2016seq,kang2017object,kang2018t,liu2018mobile,zhu2017flow,zhu2017deep,wang2018fully} have attempted to use temporal information to enhance single-image object detectors~\cite{ren2015faster} for multi-class video object detection.
There are two main types of approaches.First, temporal information can be used to refine the detections output by the detector as a post-processing step.
For example, Seq-NMS~\cite{han2016seq} re-scores detections using highly overlapping detections from surrounding frames.
Some approaches~\cite{kang2017object,kang2018t} are based on the concept of \emph{tubelet}, i.e. spatio-temporal bounding boxes that span consecutive frames.
T-CNN~\cite{kang2018t} uses tubelets, generated by tracking high confidence detections across frames, to re-score detections and recover false negatives.

The second type of approaches introduces temporal coherence while learning the features used by the model in an end-to-end manner.
FGFA~\cite{zhu2017flow} uses optical flow to estimate the motion between frames, which is employed to learn features that aggregate information from surrounding frames, while~\cite{zhu2017deep} uses it for efficiency reasons, extracting features only for selected frames and propagating them to nearby frames.
Contrary to the pixel-level approaches, Motion-Aware network~\cite{wang2018fully} introduces instance-level feature aggregation by estimating the movement of proposals across frames and combining them.
All these approaches use temporal information to improve object detection in videos, whereas we exploit it to select sets of samples in the context of active learning.

\begin{figure*}[h]
    \centering
    \includegraphics[width=\textwidth]{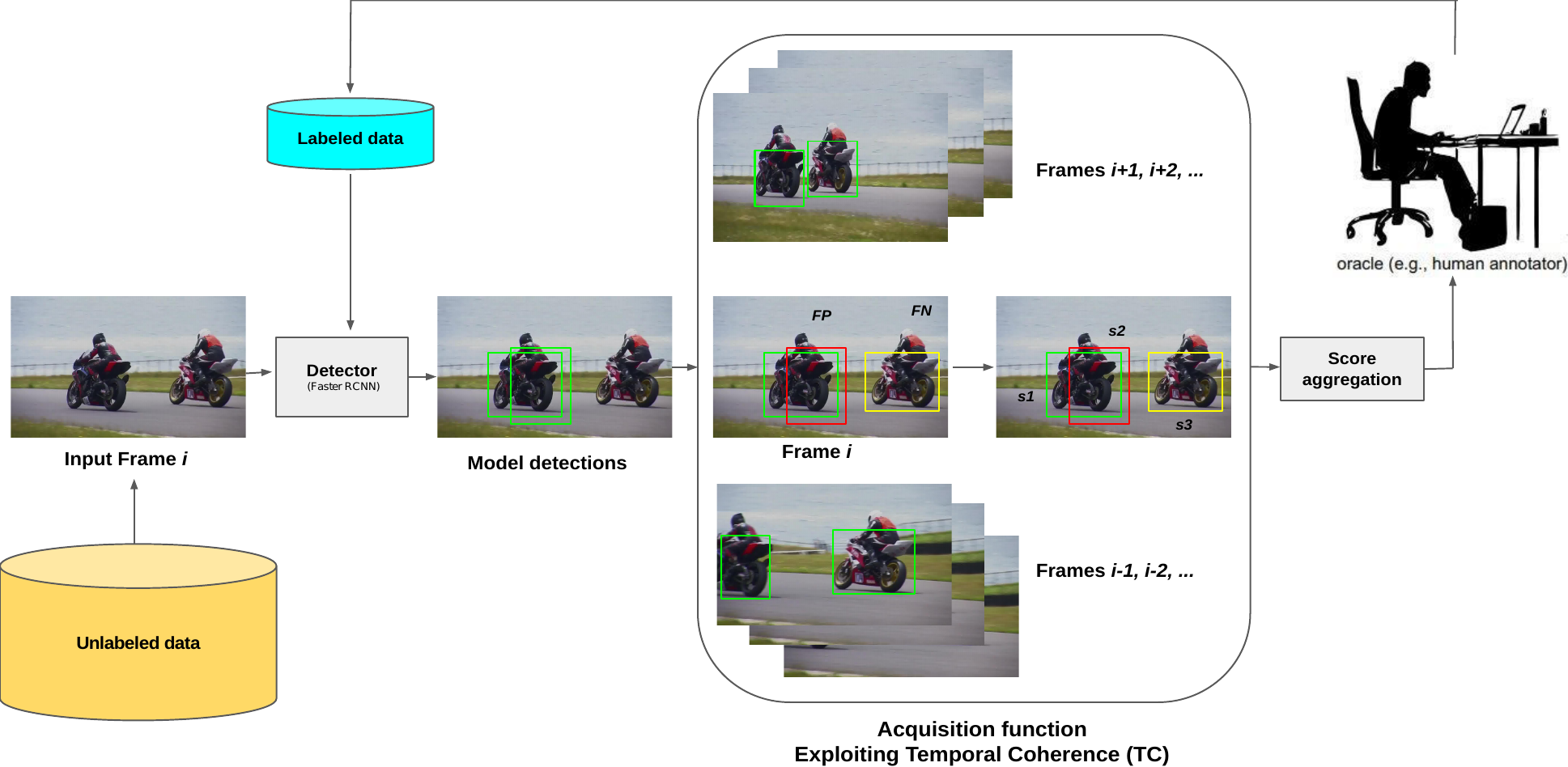}
    \caption{\small \textbf{Overview of our active learning framework exploiting temporal coherence.} The detector outputs detections (green) for each frame in the unlabeled data. Considering the relationships between the detections of neighboring frames (both forward and backward), our temporal coherence acquisition function predicts false positive (red) and false negative (yellow) errors. Based on these predictions, each frame is given an aggregated score and ranked for selection. Finally the frames with top scores are annotated and added to the labeled data.}
    \label{fig:ADL_TC_framework}
\end{figure*}

\section{Active Learning for Video Object Detection}

We describe here the general process of active learning applied to video object detection. 
Given a large pool of unlabeled data $\mathcal{D}_U$ (video frames) and an annotation budget $b$, the goal of active learning is to select a subset of $b$ samples to be annotated as to maximize the performance of an object detection model (e.g.\ Faster R-CNN~\cite{ren2015faster}). 
Active learning methods generally proceed sequentially by splitting the budget in several \emph{cycles}.
Here we consider the batch-mode variant~\cite{settles2009active}, which annotates multiple samples per cycle, since this is the only feasible option for CNN training.
At the beginning of each cycle, the model is trained on the labeled set of samples $\mathcal{D}_L$\footnote{Most methods start with a small initial labeled set selected at random.}.
After training, the model is used to select a new set of samples to be annotated at the end of the cycle via an \emph{acquisition function}. 
The selected samples are added to the labeled set $\mathcal{D}_L$ for the next cycle and the process is repeated until the annotation budget $b$ is spent.
Fig.~\ref{fig:ADL_TC_framework} presents the active learning framework with our temporal coherence acquisition function, described in sec.~\ref{sec:tc}.
Note how each sample corresponds to an entire frame and thus all objects in the frame are annotated simultaneously.

The acquisition function is the most crucial component and the main difference between active learning methods in the literature. 
In general, an acquisition function $\varphi$ receives a sample $x$ and outputs a score $\varphi(x)$ indicating how valuable $x$ is for training the current model.
More sophisticated acquisition functions may consider additional data such as the samples already selected for the current batch, the previously labeled samples $\mathcal{D}_L$, or the unlabeled pool $\mathcal{D}_U$ (see~\cite{settles2009active} for details).
In the remainder of this section, we introduce our two proposed acquisition functions for video object detection in road scenes.
Sec.~\ref{sec:errorbased} presents an exploratory function that approximates a performance upper bound.
Sec.~\ref{sec:tc} describes our main contribution: a practical acquisition function based on temporal coherence and specialized for video object detection. 

\begin{figure*}
    \centering
    \includegraphics[width=\textwidth]{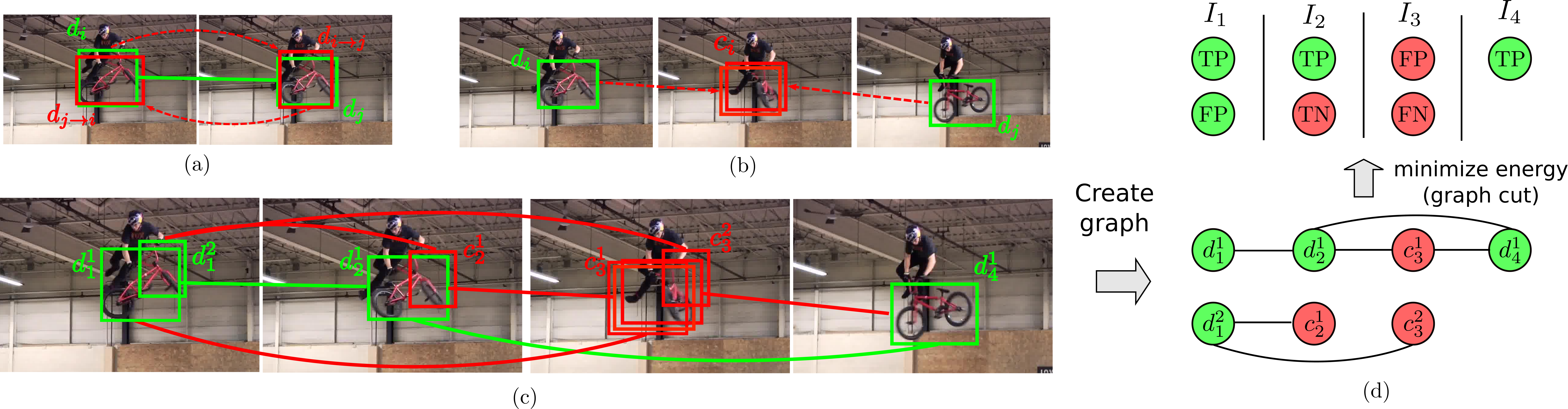}
    \caption{\small \textbf{Error estimation using temporal coherence.} (a) Detections (green) across different frames are linked depending on the overlap with their corresponding tracks (red). (b) Candidate detections (red) are obtained by clustering tracked detections that do not overlap any local detection. (c) Example of detections, candidates, and their links for four consecutive frames. (d) Nodes of the generated graph using detections and candidates corresponding to figure (c). Once the graph is created, we minimize its energy via graph-cut to obtain and estimation of the errors in terms of FP and FN. In this example, we only track up to two surrounding frames, but in practice we use three. \vspace{-2mm}}
    \label{fig:TC_method}
\end{figure*}

\subsection{Oracle-based acquisition}
\label{sec:errorbased}
The underlying assumption of active learning is that some data samples provide more valuable information than others, so that when labeled and used for training, they improve the model performance by decreasing the number of errors.
A suitable acquisition function would select those samples in which the network commits the greatest number of errors so they can be remedied. Assuming perfect generalization from training to test data, such function would be an upper bound for all active learning methods.\footnote{In practice, a decrease in errors in the training set may not necessarily lead to better performance in a separate test set, making this acquisition function an \emph{approximation} to the upper bound.}  
Motivated by this and in order to study the potential of active learning for video object detection, we propose here an \emph{oracle-based} acquisition function to implement this desirable behavior.

Our oracle-based active selection uses ground-truth information to quantify the number of errors in a given image, and selects those images that have the greatest number of errors.
Note this is not a useful active learning function in practice, as we would not have access to the ground-truth annotations in a real scenario.
We consider two types of errors that directly affect the usually employed object detection metric of Average Precision (AP)~\cite{everingham2010pascal,lin2014microsoft}: False Positives (FP) and False Negatives (FP).
Let us consider a detection as \emph{correct} if it overlaps a ground-truth bounding box more than 0.5, using the Intersection-over-Union (IoU) measure for overlap~\cite{everingham2010pascal}.
FPs are detections that are not correct (i.e.\ have little or no overlap with any ground-truth) or are duplicated, while FNs are those ground-truth instances that have not been detected.
We consider two different acquisition functions, one which considers the number of FPs in a frame and the other which considers the number of FNs in a frame\footnote{We experimented with combining both FP and FN in the acquisition function but found this to not improve results.}. 
Since the acquisition scores of these functions are integer numbers, it is frequent to have ties between images. We disambiguate between ties by random selection.
\subsection{Temporal coherence for error estimation}
\label{sec:tc}

Video data has the inherent property of \emph{temporal coherence}, i.e.\ nearby frames are expected to contain the same instances in nearby locations. 
Based on this, we propose a method to estimate the errors of a video object detector by exploiting the expected temporal coherence, and then use the estimates with the oracle-based acquisition function proposed in sec.~\ref{sec:errorbased}, but using estimations as oracle.
Let us consider a video $v$ composed of a sequence of $L$ frames $\{I_1,...,I_L\}$.
An object detector outputs a set of detections $D_i = \{d_i^0,...,d_i^K\}$ for each frame $I_{i}$\footnote{Here we consider object detectors that process each frame independently, such  as Faster R-CNN~\cite{ren2015faster}.}.
Temporal coherence induces a bijective mapping between sets of detections in nearby frames when corrected for minor localization changes.
In order to correct such changes we employ an object tracker, of which details follow later. Formally, given a detection $d_i^k$ in frame $I_i$, the tracker estimates the location of the contents of this region in frame $I_j$, which we refer to as $d_{i\to j}^k$.
The tracking can be performed in the direction of time ($i<j$) or in the reverse direction.
The set of all tracked detections $D_{i\to j}=\{d_{i\to j}\}$ can be thought of as weak detections obtained via temporal coherence using another frame's detections, rather than being directly predicted by the object detector based on the frame's content. We can now link detections of the same class across frames based on their tracked detections. More concretely, we link detection $d_i^{k}$ in frame $I_i$ with detection $d_j^{l}$ in $I_j$ if  $\IoU(d_i^{k},d_{j\to i}^{l}) > \theta$ or $\IoU(d_j^{l},d_{i\to j}^{k}) > \theta$ (Fig.~\ref{fig:TC_method}a).
That is, if any of the tracked detections (forward or backward) overlaps the other detection in the corresponding frame.
Note how there might be tracked detections that are not matched with any local detection (Fig.~\ref{fig:TC_method}b).
Such tracked detections could indicate the presence of an instance in that frame that has been missed by the detector.
We cluster groups of unmatched tracked detections in the same frame based on their overlap. 
We term these groups as detection \emph{candidates} and use the notation $c^k_i$ for the $k$-th candidate of frame $I_i$.
Each detection $d_i$ can either be a True Positive (TP) if it correctly localizes an object instance in the image, or a FP if it erroneously predicts the presence of a particular object.
On the other hand, a detection candidate $c_i$ can be a True Negative (TN) if no object instance is present in its location, or a FN if it corresponds to a missed detection.
We now estimate the type of every detection and detection candidate by formalizing our approach as a graphical model.

\vspace{-2mm}
\paragraph{Graphical model.}
Let us express all detections and candidates as a set of binary random variables $\mathcal{V}=\{v_1,...,v_N\}$, where $v_n =d$ if it corresponds to a detection $d_i^k$ and $v_n = c$ for a candidate $c_i^k$. 
Let $\mathcal{G} = (\mathcal{V},\mathcal{E})$ be an undirected graph  with vertices $\mathcal{V}$ and edges $\mathcal{E}$ between connected detections across different frames (via the links previously introduced) and candidates connected with their originating detections (see Fig.~\ref{fig:TC_method}).
Each $v_n$ can take one of four possible labels: TP, FP, TN, or FN. We consider the following energy function on label assignment $\mathcal{L}$:
\vspace{-1mm}
\begin{equation}
\label{eq:energy}
    E(\mathcal{L}) = \sum_{v \in \mathcal{V}} \phi_v(l_v) + \sum_{v_1,v_2\in \mathcal{C}} \psi_{v_1,v_2}(l_{v_1},l_{v_2}),
\end{equation}
where $\phi_v(l_v)$ is the unary cost of assigning label $l_v$ to $v$ and $\psi_{v_1,v_2}(l_{v_1},l_{v_2})$ is the pairwise cost of assigning the label pair $(l_{v_1},l_{v_2})$ to a pair of connected variables $(v_1,v_2)\in \mathcal{E}$.
We define the unary cost for detection variables as
\vspace{-2mm}
\begin{equation}
\vspace{-2mm}
    \phi_{v=d}(l_v) = 
    \begin{cases}
    0 & \text{if $l_v=$ TP}\\
    \infty & \text{if $l_v=$ TN}\\
    1 & \text{if $l_v=$ FP}\\
    \infty & \text{if $l_v=$ FN}\\
    \end{cases}
\end{equation}
This indicates that in principle we trust the outputs of the detector and that assigning a contradicting label should incur some cost. By definition, detections are `positives' and thus assigning a `negative' label is strongly discouraged.
Analogously, the unary cost for candidate variables is 
\vspace{-2mm}
\begin{equation}
\vspace{-2mm}
    \phi_{v=c}(l_v) = 
    \begin{cases}
    \infty & \text{if $l_v=$ TP}\\
    0 & \text{if $l_v=$ TN}\\
    \infty & \text{if $l_v=$ FP}\\
    1 & \text{if $l_v=$ FN}\\
    \end{cases}
\end{equation}
In this case, candidates can only be negatives as they are not part of the original outputs of the detector and hence cannot be positives.

We specify the pairwise cost using the following matrix
\vspace{-2mm}
\begin{equation}
\vspace{-2mm}
\label{eq:pairwise}
   \psi_{v_1,v_2}(l_{v_1},l_{v_2})=
\begin{pmatrix}
    0 & 1 & 1 & 0\\
    1 & 0 & 0 & 1\\
    1 & 0 & 0 & 1\\
    0 & 1 & 1 & 0\\
\end{pmatrix},
\end{equation}
where the considered label assignment order is $l_{v}=$ (TP, FP, TN, FN).
This indicates that TP should be connected with other TP or FN, whereas FP are preferably connected with other FP or with TN.
Intuitively, the pairwise cost enforces temporal coherence between the detections and the candidates, propagating the correctness to connected variables and collaboratively determining the errors.

We optimize the energy function in~\eqref{eq:energy} via graph cut~\cite{kolmogorov2004energy}, which finds the globally optimal solution by solving the dual max-flow problem.
In fact, the problem can be reduced to a binary labelling problem, considering only two possible labels (True or False) with different meanings depending on the type of input variable, i.e. positives for detections and negatives for candidates.
We use the graph-cut implementation in the Python library PyMaxflow~\cite{boykov2004experimental}. 
\paragraph{Acquisition function.}

Once all variables in $\mathcal{V}$ have been assigned their optimal labels, we record the estimated number of FPs and FNs contained in each frame.
We revert now to the oracle-based acquisition function described in sec.~\ref{sec:errorbased}, but using error estimates instead of actual errors, which makes the function is useful in practice as it does not require any ground-truth information. 
We refer to this acquisition function as Temporal Coherence (TC). 
Experimental results show similar performance when considering only FP, only FN, or both FP and FN. Therefore, we use only the number of FP for the acquisition function of TC.
\paragraph{Object tracker.}
In order to temporally link detections and construct connections between graph nodes, we considered two types of object trackers, namely Optical Flow (PWC-NET)~\cite{Sun_2018_CVPR} and SiamFC tracker~\cite{bertinetto2016fully}. 
To utilize optical flow for the purpose of object tracking, we first compute a dense 2D real-valued vector map of the motions between all pairs of consecutive frames in the dataset. 
Then, we translate the box coordinates using the motion vector corresponding to the box center to obtain the tracked box in the next or previous frame.
As an alternative to track detections we employ SiamFC~\cite{bertinetto2016fully}, a state of the art Siamese-based object tracker. 
The bottleneck of this tracking method in the context of active learning is that, despite its efficiency, it imposes a huge computational burden when tracking detections every cycle, given the vast amount of detections. 
On the contrary, optical flow is only computed once at the beginning and can be used throughout all cycles with a negligible overhead.
\vspace{-2mm}
\section{Synthetic Dataset}
\label{sec:dataset}
\vspace{-2mm}

\begin{table}[t]
    \centering
    \resizebox{\columnwidth}{!}{
    \begin{tabular}{l ccccc}
        \toprule
         Subset Name & Seq. & Frames & Area & Conditions & P(Pe/Cy/Ca/Wh)\\
         \midrule
         Default & 150 & 74K & C,H & S,W,F,R & 30/20/35/0 \\
         Town & 36 & 17K &  T & S,W,F,R & 30/20/35/0 \\
         Night & 6 & 3K & C,H & N & 0/0/35/0 \\
         Wheelchair & 5 & 2K & C,T & S & 20/20/0/100 \\
          Test (no WC) & 85 & 40K & C,H,T & S,F,R,N & 30/20/35/0 \\
          Test (WC) & 12 & 5K & C,T & S & 20/20/0/100 \\
         \bottomrule
    \end{tabular}}
    \vspace{2mm}
    \caption{\small \textbf{SYNTHIA-AL data distribution.} Seq. indicates the number of videos. Environment conditions are Fall (F), Winter (W), Spring (S), Rain (R), and Night (N). Areas are City (C), Town (T), and Highway (H). The spawning probabilities are given for pedestrians (Pe), cyclists (Cy), cars (Ca), and wheelchairs (Wh). \vspace{-4mm}}
    \label{tab:synthetic_data_distribution}
\end{table}

Most active learning methods~\cite{gal2017icml,sener2018active,settles2009active} are evaluated on simple image classification datasets such as MNIST~\cite{lecun1998mnist} or CIFAR~\cite{krizhevsky2009learning}.  
Approaches specific for object detection~\cite{brust2019active,roy2018bmvc,grauman2014ijcv,Yoo_2019_CVPR} mainly use PASCAL VOC~\cite{everingham2010pascal}, covering various scene types.
In the context of autonomous driving, only~\cite{roy2018bmvc} uses a dataset depicting road scenes, KITTI~\cite{geiger2013vision}.
Similarly to several other image datasets for autonomous driving~\cite{Cordts2016Cityscapes,yu2018bdd100k}, KITTI is manually curated to mostly contain relevant knowledge usable to train object detection models.
This process is performed by human annotators that select interesting data samples containing cars, pedestrians, etc.
The goal of active learning, however, is automatizing this process, making existing datasets not suitable for a proper evaluation.
Ideally, a good dataset for evaluating active learning contains a more raw version of the data, in which the image distribution is unbalanced towards the uninteresting (e.g. empty road scenes) and highly redundant.
Such dataset would better represent the type of data collected in a real setting, for example, video captured from a driving car.
For this reason, and following recent trends~\cite{Ros:2016,Richter:2017},  we have created a new synthetic dataset to evaluate active learning for object detection in road scenes. In particular, we modified the SYNTHIA environment~\cite{Ros:2016} to generate the SYNTHIA-AL dataset\footnote{Available at \url{http://www.synthia-dataset.net}} using Unity Pro game engine. The aim is having an unbalanced foreground/background distribution, simulating the real collection scenario of a driving car.
Moreover, a set of object classes and conditions should be predominantly present, while other classes and conditions must appear less frequent.
The data is generated by driving a car in a virtual world consisting of three different areas, namely town, city, and highway. These areas are populated with a variety of pedestrians, cars, cyclists, and wheelchairs, except for the highway which is limited to cars. 
These dynamic objects are arbitrarily spawned at predefined positions with a given probability and follow randomly predefined paths without leaving each area. 
Several environmental conditions can be set: season (winter, fall, spring), day time (day or night), and weather (clear or rainy).
By default, we always use spring and clear during the day, and only change one condition at a time. 
Objects with no lights can be hard to visualize during the night, so we only use cars for the night condition.
Figure~\ref{fig:qual_results} shows examples of images in the dataset.

Table~\ref{tab:synthetic_data_distribution} provides the specification of the dataset.
The video sequences are captured at 25 fps with a random length between 10 and 30 seconds. 
We have generated one subset with the default parameters and three smaller subsets with altered conditions. 
The first subset consists of 150 sequences, which amounts to 75\% of all the data, with the default settings, i.e. containing cars, pedestrians, and cyclists, under different daily conditions, but only in the city and highway areas.
The second subset contains 36 sequences (20\% of the dataset) captured in the town area instead. 
The night condition only represents 3\% of the whole data (6 sequences) and it is fully contained in the third subset. 
Finally, we have added wheelchairs and removed cars in the fourth subset, which represents the 2\% of the dataset with only 5 sequences. 
The test set contains 85 sequences with balanced distributions on areas and conditions (except winter) on the three main classes plus another 12 sequences including wheelchairs. All images are automatically annotated with 2D bounding boxes and class labels for every object that can be reasonably seen (more than 50 pixels). 

\vspace{-1mm}
\section{Experimental Setup}
\subsection{Active learning procedure}
\vspace{-2mm}

All considered active learning methods follow the same procedure and employ the same state-of-the-art object detector based on Faster R-CNN~\cite{ren2015faster}.
We start with the model pre-trained on COCO~\cite{lin2014microsoft}, which contains 80K images from 80 different object categories.
The initial labeled set $\mathcal{D}_L$ consists of $2\%$ of train dataset that is selected randomly once for all the methods. 
At each cycle, we fine-tune the latest model of the previous cycle, as we have experimentally observed that this leads to faster convergence than fine-tuning the initial model or from scratch as in ~\cite{chitta2018large}. We have also seen that in order not to get stuck in local minima, the learning rate should be high enough. 
Once the new model is fine-tuned, we use it with the corresponding acquisition function to select $b/C$ frames, which are then labeled and added to $\mathcal{D}_L$. 
We continue for $C$ cycles until budget $b$ is completely exhausted.
In all experiments, the budget per cycle is 2\% of the dataset.
\vspace{-4mm}
\paragraph{Evaluation.}
For each cycle, we evaluate the model trained with the updated labeled set for that cycle on the test set. Detections are processed using Non-Maxima Supression~\cite{felzenszwalb2010object} and thresholded by score, rejecting all detections below 0.5.
We use AP averaged over all classes using a detection threshold of IoU$>0.5$. 
\vspace{-4mm}
\paragraph{Implementation details.}
We used Tensorflow's Object Detection API~\cite{huang2017speed} as the base code to develop our experiments. 
We trained all models with the momentum optimizer with value 0.9 and the initial learning rates 0.02 and 0.001 for SYNTHIA-AL and ImageNet-VID~\cite{ImageNetVID2015} datasets, respectively. 
We train for 10 epochs and reduce the learning rate by a factor of 5 once after 5 epochs and again at 7 epochs for SYNTHIA-AL. 
In the case of ImageNet-VID we reduce the learning rate at epochs 3 and 5, training a total of 6 epochs.
For efficiency reasons, we resize all images to fixed height of 300 pixels and preserve the aspect ratio. We use a batch size of 12 for all the experiments. Finally, to obtain more stable results we repeat the experiments 3 times and report the mean and standard deviation in our results.
\subsection{Baselines}
\label{sec:baselines}

\vspace{-2mm}
\paragraph{Random.}
Random sampling  selects an arbitrary subset of frames from all unlabeled frames. Given the extreme imbalance inherent to video data due to varying video length, uniform random sampling selects most frames from the longer videos while under-representing shorter videos, which damages the performance.
Moreover, video data is redundant due to the high similarity between nearby frames, which makes annotating the surrounding frames of an already annotated frame wasteful.
For these reasons, we also consider an improved random sampling procedure that includes temporal representativeness,  which prevents selecting the $k$ neighbors in both directions of already labeled frames. In the experiments, we set the $k$ to 3 for ImageNet-VID dataset and 1 for SYNTHIA-AL dataset for all the methods.
This criterion naturally increases the diversity of the selected batches at each cycle by limiting the similarity between data samples. 
We call this baseline \emph{Random+R}.
\begin{figure*}[t]
\centering
\includegraphics[width=\textwidth]{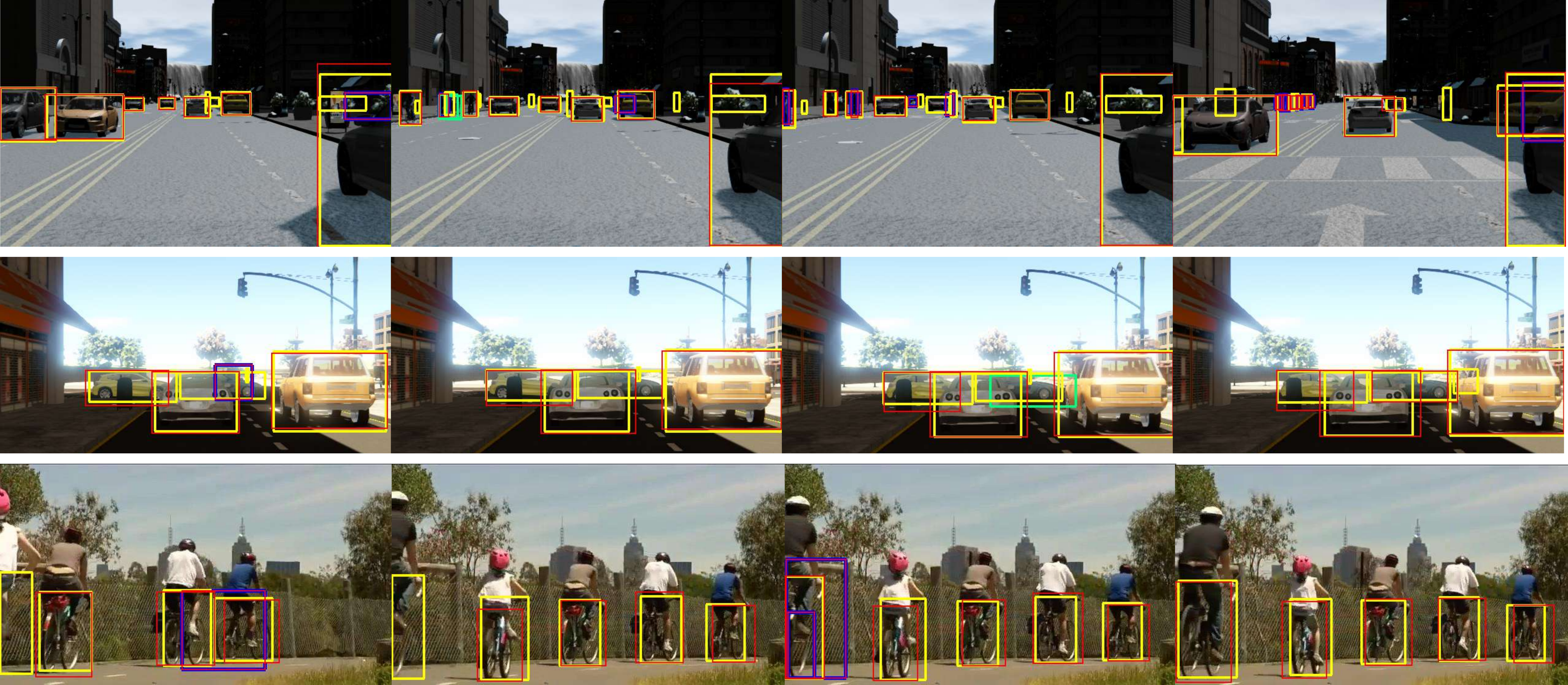}
\caption{\small \textbf{Examples of errors detected by our temporal coherence approach on SYNTHIA-AL (top, middle) and ImageNet-VID~\cite{ImageNetVID2015} (bottom)}. We show ground-truth boxes in yellow and output detections in red. After solving our graphical model based on temporal coherence, some of the detections are considered as false positives (purple), while other boxes are added as false negatives (green).\vspace{-4mm}}
\label{fig:qual_results}
\end{figure*}
\vspace{-2mm}
\paragraph{Uncertainty.}
We consider three other baselines based on uncertainty measures used in recent active learning approaches for object detection~\cite{brust2019active,roy2018bmvc}.
\emph{Least confidence}~\cite{lewis1994sequential,roy2018bmvc} considers the score of the most probable class and selects those samples that have the lowest score on it.
\emph{Entropy}~\cite{dagan1995committee} is an information theory measure that captures the average amount of information contained in the predictive distribution, attaining its maximum value when all classes are equally probable.
In both cases, we use the average score of all detections in the image to obtain a single score per image.
\emph{Margin sampling}~\cite{settles2009active,brust2019active} uses the difference between the two classes
with the highest scores as a measure of proximity to the decision boundary. Following~\cite{brust2019active}, we sum all margin sampling scores of individual detections to aggregate them into an overall image score. 
\subsection{Datasets}
\vspace{-2mm}
Besides our SYNTHIA-AL dataset (sec.~\ref{sec:dataset}), we also perform experiments on a real-image dataset, ImageNet-VID~\cite{ImageNetVID2015}, which is commonly used as video object detection benchmark. 
Since the focus of this paper is video object detection in road scenes, we select 3 classes that are likely to be encountered in the context of autonomous driving, namely: car, bike, and motorcycle.
Selecting all videos that contain these classes amounts to 795 videos in the training set and 87 videos in the validation set, which we use for test. 
The length of the videos varies between a few frames to over 1000.
We have cleaned this dataset by manually discarding all those frames that had missing annotations, which amounts to 20K frames in the training set and 5K frames in the validation set.
The final dataset contains 129K frames for training and 14K frames for validation. 
\begin{figure*}[ht]
\centering
\includegraphics[width=\textwidth]{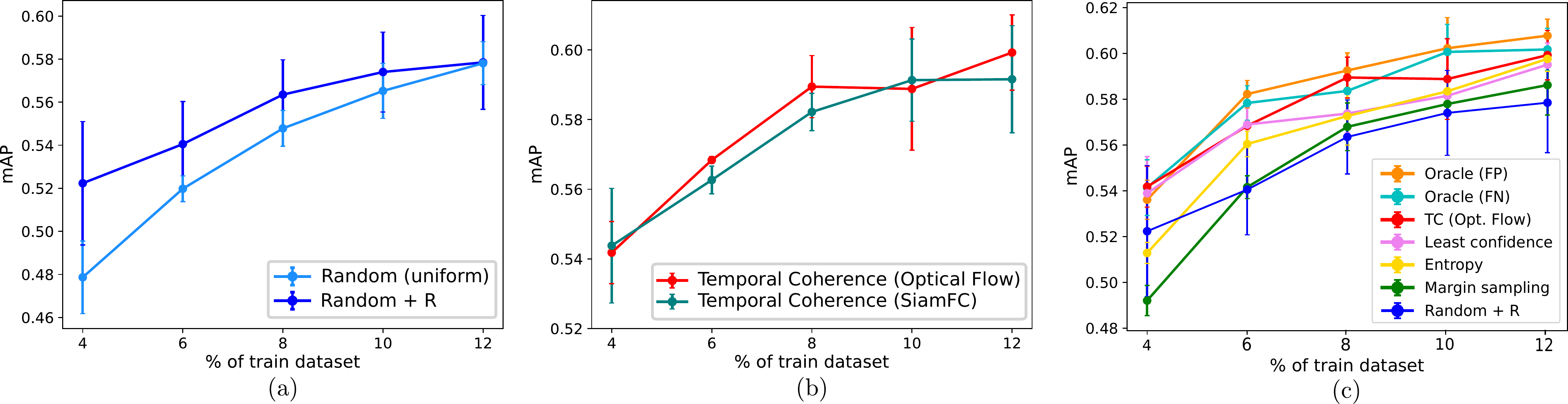}
\caption{\small \textbf{Results on SYNTHIA-AL.} (a) Random baselines with and without representativeness. (b) Our Temporal Coherence using either Optical Flow or SiamFC. (c) Baselines, oracle-based acquisition, and Temporal Coherence. All curves are the average of 3 runs. \vspace{-2mm}}
\label{fig:results_synthia}
\end{figure*}

\section{Results} 
\vspace{-2mm}
We present active learning results using performance (mAP) curves as a function of the number of selected samples, as usually reported in the literature~\cite{gal2017icml,sener2018active}.
This allows us assess the benefit of each active learning method for different total number of samples used to train the model. 
For each method, we plot the average performance for all runs with vertical bars to represent the standard deviation.
 
We first validate the ability of our graphical model (sec.~\ref{sec:tc}) to estimate detection errors using temporal coherence.
Fig.~\ref{fig:qual_results} presents some resulting predictions on both datasets.
We can see how many FP (purple) are correctly detected, including those corresponding to double detections (top row, rightmost column).
Moreover, FN (green) are discovered due to the forward and backward tracking of surrounding detections (middle row, third column).

\vspace{-1mm}
\subsection{SYNTHIA-AL}
\vspace{-2mm}

Fig.~\ref{fig:results_synthia} presents all quantitative results on our SYNTHIA-AL dataset.
We start by evaluating the difference between the two random baselines: uniform and our enhanced Random+R baseline (Fig.~\ref{fig:results_synthia}a). 
We can observe how the addition of representativeness is clearly beneficial for active learning in video object detection.
In the remainder of the paper, we always include temporal representativeness and per-video sampling for all evaluated methods.

\begin{table}[t]
    \centering
    \resizebox{\columnwidth}{!}{
    \begin{tabular}{l cccc}
        \toprule
         \multirow{2}{*}{Methods} & \multicolumn{2}{l}{SYNTHIA-AL} &  \multicolumn{2}{c}{ImageNet-VID} \\
         & mAP & Rel. & mAP & Rel. \\
         \midrule
         All data & 0.628 & 100\% & 0.839 & 100\% \\
         \midrule
         Random+R & 0.578 & 92.0\% & 0.821 & 97.8\%\\
         Least Confidence & 0.595 & 94.7\%& 0.818 & 97.4\%\\
         Margin sampling & 0.586 & 93.3\% & 0.820 & 97.7\%\\
         Entropy & 0.597 & 95.0\% & 0.821 & 97.8\%\\
         \midrule
         Oracle (FP) & 0.607 & 96.6\% & - & - \\
         Oracle (FN) & 0.601 & 95.7\%& - & -\\         
         Temporal Coherence (SiamFC) & 0.591 & 94.1\% & - & - \\ 
         Temporal Coherence (Opt. Flow) & \textbf{0.599} & \textbf{95.3\%} & \textbf{0.830} & \textbf{98.9\%}\\
         \bottomrule
    \end{tabular}}
    \vspace{2mm}
    \caption{\small  \textbf{Active learning results.} The first row shows the performance (mAP) obtained when using the entirety of the dataset. All other rows show the performance of all methods using $12\%$ of all data in SYNTHIA-AL and $10\%$ of ImageNet-VID~\cite{ImageNetVID2015}, both in absolute performance and relative to using all data. \vspace{-4mm}}
    \label{tab:performance_wrt_full_set}
\end{table}
Next, we evaluate the effect of the two types of trackers considered in our temporal coherence method, SiamFC~\cite{bertinetto2016fully} and Optical Flow~\cite{Sun_2018_CVPR}, within the active learning cycles.
Fig.~\ref{fig:results_synthia}b presents the quantitative evaluation of temporal coherence with either tracker. The results show that there is no improvement gained by using the more sophisticated SiamFC tracker compared to Optical Flow. 
Furthermore, Optical Flow can significantly speed up the active learning process.
In this case, the motion vectors are computed once at the beginning of the process, whereas SiamFC needs to perform expensive computations at every cycle. 
Finally, we compare Temporal Coherence (TC) with all baselines. 
To explore an upper bound for TC, we also consider the oracle-based methods of section~\ref{sec:errorbased}, selecting those frames with the highest number of FP or FN based on ground-truth information.
These methods are designated by Oracle (FP) and Oracle (FN), respectively.
The results in Fig.~\ref{fig:results_synthia}c show that our TC method outperforms all three uncertainty based methods and the random baseline. 
The narrow gap between our TC method and the oracle-based methods implies that FP and FN predictions of the graphical model are effective estimates of the actual errors that the model can learn from. 
Moreover, TC enables us to achieve more than $95\%$ of performance of the model trained on entire dataset by annotating only $12\%$ of the data. Table~\ref{tab:performance_wrt_full_set} shows the effectiveness of active learning methods in videos by using a small portion of datasets.

\vspace{-2mm}
\subsection{ImageNet-VID}
\vspace{-2mm}
\begin{figure}[t]
\centering
\includegraphics[width=.85\columnwidth]{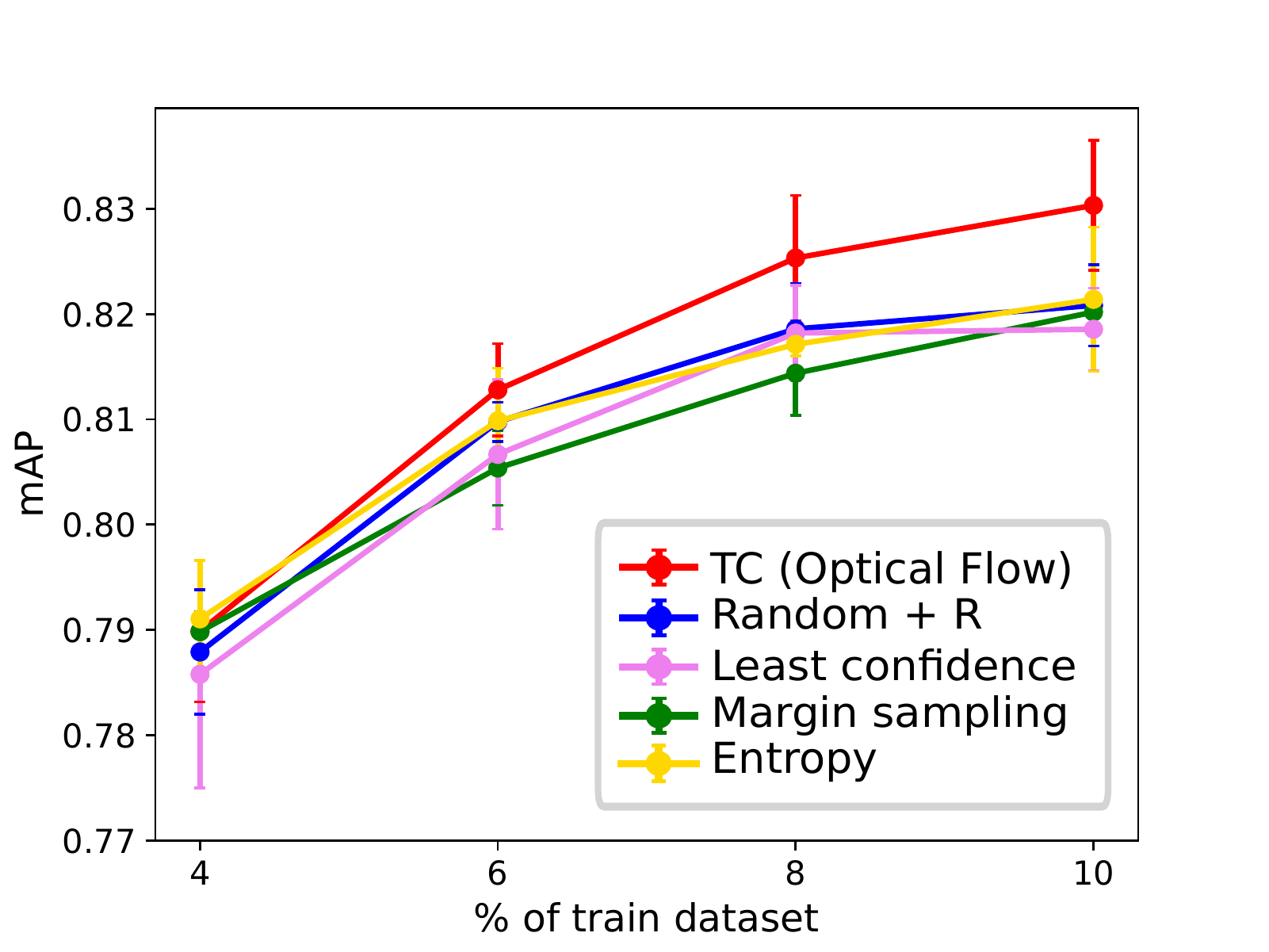}
\caption{\small \textbf{Results on ImageNet-VID~\cite{ImageNetVID2015}.} Average of 3 runs. \vspace{-6mm}}
\label{fig:imagenet_baselines}
\end{figure}

 To evaluate our temporal coherence method on a dataset of real images, we perform experiments on ImageNet-VID~\cite{ImageNetVID2015}. 
 Fig.~\ref{fig:imagenet_baselines} compares TC with Optical Flow against uncertainty and random baselines. 
 The results illustrate that TC is superior to all the baselines for all cycles. Additionally, Table~\ref{tab:performance_wrt_full_set} shows that TC manages to attain almost the full performance of a model trained with the entire dataset by using only $10\%$ of the data, which is a significant reduction in the annotation effort.

\vspace{-2mm}
\section{Conclusions}
\vspace{-2mm}
In this paper, we introduced a novel active learning approach for object detection in videos which exploits the temporal coherence. Our approach is formulated in terms of an energy minimization function of a graphical model built on tracked object detections. Additionally, we introduced a new synthetic dataset specially designed to evaluate active learning for object detection in the context of autonomous driving. Experimental results conducted on two datasets showed that our approach outperformed major active learning baselines. A drawback of temporal coherence based active learning is that it is computationally more demanding than the baselines. We plan to minimize the computational overhead of our system in future research.

	\begin{small}
	\noindent\textbf{Acknowledgements.} The authors thank Audi Electronics Venture GmbH for their support during the development of this work, the Generalitat de Catalunya CERCA Program and its ACCIO agency,  Unity for the support in the synthetic dataset generation, the EU Project CybSpeed MSCA-RISE-2017-777720 and CYTED Network (Ref. 518RT0559). Antonio thanks the financial support by ICREA under the ICREA Academia Program, and the Spanish project TIN2017-88709-R (MINECO/AEI/FEDER, UE).
	\end{small}
	
{\small
\bibliographystyle{ieee}
\bibliography{shortstrings,egbib}
}

\end{document}